\documentclass[sigconf]{acmart}
\usepackage{kotex}
\usepackage{caption}
\usepackage[ruled,vlined]{algorithm2e}
\usepackage{adjustbox}
\usepackage{cite}
\usepackage{colortbl}
\usepackage{amsmath,amsfonts}
\usepackage{algorithmic}
\usepackage{graphicx}
\usepackage{textcomp}
\usepackage{xcolor}
\usepackage{enumitem}
\def\BibTeX{{\rm B\kern-.05em{\sc i\kern-.025em b}\kern-.08em
    T\kern-.1667em\lower.7ex\hbox{E}\kern-.125emX}}
\usepackage[ruled,vlined]{algorithm2e}
\usepackage{booktabs}

\AtBeginDocument{%
  \providecommand\BibTeX{{%
    Bib\TeX}}}

\setcopyright{acmlicensed}
\copyrightyear{2026}
\acmYear{2026}
\setcopyright{cc}
\setcctype{by}
\acmConference[WWW '26]{Proceedings of the ACM Web Conference 2026}{April 13--17, 2026}{Dubai, United Arab Emirates}
\acmBooktitle{Proceedings of the ACM Web Conference 2026 (WWW '26), April 13--17, 2026, Dubai, United Arab Emirates}
\acmPrice{}
\acmDOI{10.1145/3774904.3792915}
\acmISBN{979-8-4007-2307-0/2026/04}

\begin{document}

\title{Back to the Future: Look-ahead Augmentation and Parallel Self-Refinement for Time Series Forecasting}

\author{Sunho Kim}
\affiliation{%
  \institution{Korea University}
  \department{Computer Science and Engineering}
  \city{Seoul}
  \country{Republic of Korea}
}
\email{sunho_kim@korea.ac.kr}

\author{Susik Yoon}
\affiliation{%
  \institution{Korea University}
  \department{Computer Science and Engineering}
  \city{Seoul}
  \country{Republic of Korea}
}
\email{susik@korea.ac.kr}

\begin{abstract}

Long-term time series forecasting (LTSF) remains challenging due to the trade-off between parallel efficiency and sequential modeling of temporal coherence.
Direct multi-step forecasting (DMS) methods enable fast, parallel prediction of all future horizons but often lose temporal consistency across steps, while iterative multi-step forecasting (IMS) preserves temporal dependencies at the cost of error accumulation and slow inference.
To bridge this gap, we propose Back to the Future (BTTF), a simple yet effective framework that enhances forecasting stability through look-ahead augmentation and self-corrective refinement.
Rather than relying on complex model architectures, BTTF revisits the fundamental forecasting process and refines a base model by ensembling the second-stage models augmented with their initial predictions.
Despite its simplicity, our approach consistently improves long-horizon accuracy and mitigates the instability of linear forecasting models, achieving accuracy gains of up to 58\% and demonstrating stable improvements even when the first-stage model is trained under suboptimal conditions.
These results suggest that leveraging model-generated forecasts as augmentation can be a simple yet powerful way to enhance long-term prediction, even without complex architectures. 
\end{abstract}



\maketitle

\suppressfloats[t] 
\section{Introduction}

\begin{figure}[t]

  \centering
\includegraphics[width=0.90\columnwidth]{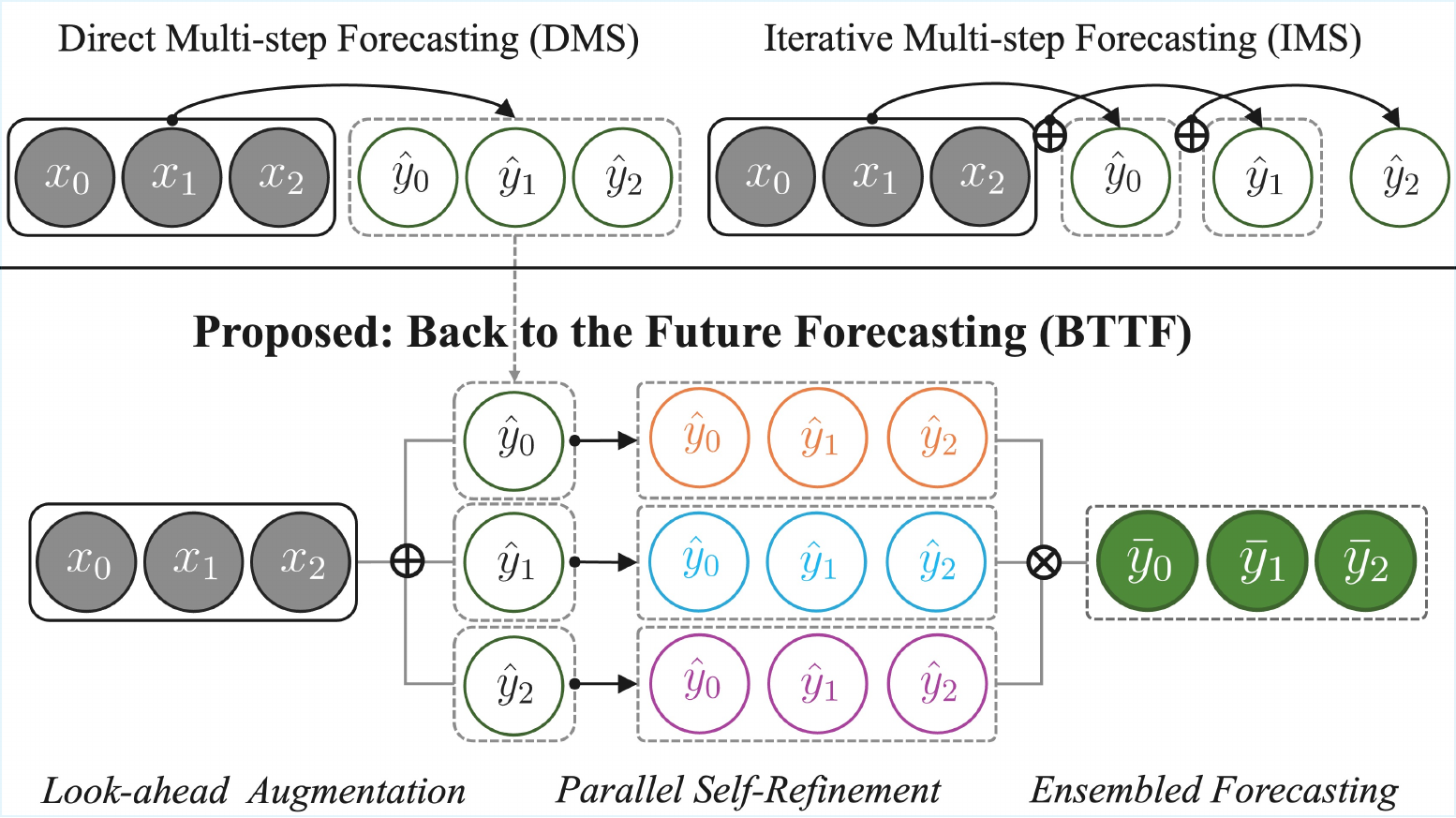} 
  \vspace{-0.3cm}
  \caption{BTTF effectively leverages the strengths of both DMS and IMS for long-term time series forecasting.}
  \vspace{-0.7cm}
  \label{fig:motivation}
\end{figure}

Time series forecasting is essential in various domains, including finance, transportation, and Web-scale services. As Internet systems continue to scale over time, accurate long-term time series forecasting (LTSF) has become crucial for reliability and adaptive decision-making in real-world applications~\citep{rest, traffic, bigforecasting}. There are two predominant strategies for LTSF: \textit{direct multi-step forecasting} (DMS), which predicts all future steps in parallel but may struggle to capture temporal dependencies, and \textit{iterative multi-step forecasting} (IMS), which preserves temporal causality via autoregressive prediction at the cost of higher computation and error accumulation.

Most existing methods attempt to mitigate these limitations and better exploit historical inputs mainly through architectural enhancements, such as hierarchical recurrent units~\citep{segrnn} and self-attention mechanisms~\citep{itransformer}. However, a recent study has shown that a lightweight linear model can be sufficient for modeling historical inputs, outperforming advanced methods with complex architectures~\citep{dlinear}. Inspired by this insight, we aim to leverage both DMS and IMS strategies, while exploiting the simplicity of linear models.

Specifically, we propose \textit{Back to the Future (BTTF)} for LTSF, a simple yet effective approach based on look-ahead augmentation and parallel self-refinement. As illustrated in Figure \ref{fig:motivation}, BTTF leverages segments of initial future predictions together with the historical inputs to ensemble refined forecasts. Consequently, it benefits from DMS-style parallel efficiency while implicitly maintaining the inter-temporal dependencies enabled by IMS. The merits of BTTF lie in its systematic capability to quantitatively diagnose and selectively exploit base models across diverse time horizons, thereby mitigating potentially noisy and unstable first-stage forecasts.

Our contributions are summarized as: (i) To the best of our knowledge, this is the first work to simultaneously combine IMS-like leverage of future predictions with DMS-like parallelism for LTSF; (ii) We instantiate the proposed idea on simple linear models through a systematic two-stage training scheme with look-ahead augmentation and parallel self-refinement; (iii) Experimental results demonstrate that the proposed strategy achieves up to 58\% performance improvement for linear models on the benchmark datasets, notably not only when they already outperform advanced baselines but also in cases where their first-stage training was unstable.

\section{Related Work}
Recent advances in LTSF have explored diverse architectural designs or data augmentation approaches. Notably, DLinear~\citep{dlinear} achieves strong performance with minimal model complexity by decomposing time series into trend and seasonal components. 
Others focus on modeling complicated temporal dependencies through specialized architectures. SegRNN~\citep{segrnn} operates on fixed-length segments with hierarchical recurrent units to alleviate long-term dependency issues. 
iTransformer~\citep{itransformer} treats each variable as a token and applies variable-wise attention using temporally encoded embeddings. 
Another line of research has focused on data-driven performance improvement for LTSF~\citep{wentime}. Zhang et al.~\citep{staug} explore data-level augmentation for time-series forecasting via joint time- and frequency-domain transformations to promote temporal coherence, whereas Demirel and Holz~\citep{demirel2023finding} introduce frequency-domain mixup for quasi-periodic signals with adaptive augmentation strength.

While existing models have achieved performance gains through increasingly complex architectural designs, their improvements still rely on exploiting historical inputs. Similarly, augmentation-based approaches remain limited to transformations of observed time series and thus do not explicitly incorporate future-aware signals. In contrast, our proposed strategy leverages model-generated predictions as stable and temporally consistent augmentation signals, enabling effective future-aware refinement with lightweight model architectures. This bridges the gap between architecture-centric (i.e., DMS-style) and data-centric (i.e., IMS-style) paradigms.





\section{Problem Setting}

Given a time series $\mathcal{T} = \{x_t\}_{t=1}^T$ with an index $t$ satisfying $L \le t \le T-H$, a LTSF model $f_\theta: \mathbb{R}^L \rightarrow \mathbb{R}^H$, parameterized by $\theta$, takes the most recent $L$ observations $\mathbf{x}_{\text{in}} = (x_{t-L+1}, \ldots, x_t) \in \mathbb{R}^L$ as input and outputs the next $H$ future prediction values $\hat{\mathbf{y}} = (\hat{x}_{t+1}, \ldots, \hat{x}_{t+H})$. i.e., $\hat{\mathbf{y}} = f_\theta(\mathbf{x}_{\text{in}})$. 

To predict the future sequence $\hat{\mathbf{y}}$, DMS-based models predict all $H$ steps in parallel as $\hat{\mathbf{x}}_{t+1:t+H} = f_\theta(x_{t-L+1}, \ldots, x_t)$, 
which enables efficient long-horizon inference but may weaken temporal dependency modeling across forecasted steps. 
In contrast, IMS-based models generate each future value autoregressively, $\hat{x}_{t+h} = f_\theta(\hat{x}_{t+h-L}, \ldots, \hat{x}_{t+h-1})$ for $h = 1, \ldots, H$, preserving temporal causality but suffering from error accumulation and slower inference.
\section{Methodology}

\begin{algorithm}[t]
\caption{Overall Procedure of BTTF (Section 4)}
\KwIn{Time series $\{\mathbf{x}_t\}_{t=1}^T$, Base Model $f_\theta$, Segment Count $N$, Step Increment $M$}
\KwOut{Final Output Sequence}

$f_\theta \leftarrow$ Train first-stage model
\BlankLine
/* \textbf{\texttt{Look-ahead Augmentation (Section 4.2)}} */ \\
$\hat{\mathbf{x}}^{(1)} \leftarrow$ First-stage model predictions on $\mathbf{x}_{t-L+1:t}$ \\
$\{\mathbf{x}^{(i)}_{\text{aug}}\}_{i=1}^N \leftarrow$ Split $\hat{\mathbf{x}}^{(1)}$ into segments and append to input

\BlankLine
/* \textbf{\texttt{Parallel Self-Refinement (Section 4.3)}} */ \\
$\{f^{(i)}_\phi\}_{i=1}^N \leftarrow$ Train $N$ second-stage models on $\mathbf{x}^{(i)}_{\text{aug}}$ \\
$\{\hat{\mathbf{x}}^{(2)}_i\}_{i=1}^N \leftarrow$ Second-stage model predictions on $\mathbf{x}_{t-L+1:t}$

\BlankLine
/* \textbf{\texttt{Ensembled Forecasting (Section 4.4)}} */ \\
Rank models by validation performance \\
For $K = M,2M,\ldots,N$, compute top-$K$ ensemble prediction \\
$K^* \leftarrow$ Select optimal $K^*$ by prediction variance analysis \\
\Return Final Output Sequence $\hat{\mathbf{x}}^{(K^*)}$
\end{algorithm}

\begin{figure*}[t]
  \centering
  \vspace{-2mm}
  \includegraphics[width=1.0\textwidth]{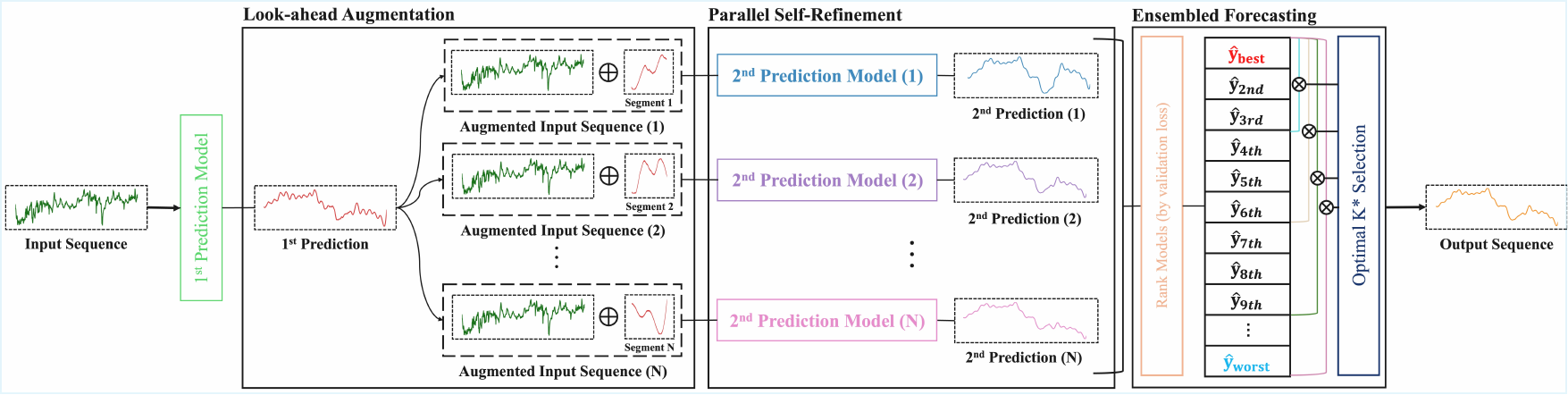}
  \vspace{-4mm}
  \caption{
Overall procedure of BTTF: 
(i) Look-ahead Augmentation, where the initial predictions are split into segments and appended to inputs to form augmented sequences; (ii) Parallel Self-Refinement, where $N$ second prediction models are trained on these augmented inputs to refine first predictions; and (iii) Ensembled Forecasting, which ranks models by validation performance, aggregates them via a step-wise top $K$ ensemble, and selects the optimal $K^*$ to generate the final output sequence.
}
  \label{fig:method}
  \vspace{-2mm}
\end{figure*}

\subsection{Overview}


Algorithm 1 and Figure 2 illustrate the proposed BTTF framework, which operates in three stages: \textbf{(i) Look-ahead Augmentation}, which augments inputs with segments of initial future predictions; \textbf{(ii) Parallel Self-Refinement}, where multiple second-stage models are trained to implicitly refine the first prediction in parallel; and \textbf{(iii) Ensembled Forecasting}, which aggregates refined predictions using a step-wise top-$K$ strategy to produce the final output.

\subsection{Look-ahead Augmentation}
Given a first-stage model trained on the original time series, we incorporate future-aware context into the input. Specifically, we create augmented inputs by attaching different portions of the initial prediction to the original sequence. Given the past window $\mathbf{x}_{t-L+1:t}$, the first-stage model 
produces an $H$-step forecast: $\hat{\mathbf{x}}^{(1)}_{t+1:t+H} = f_{\theta}\!\left(\mathbf{x}_{t-L+1:t}\right)$.
This predicted sequence serves as the source from which we extract future segments for augmentation. We divide the first-stage prediction into $N$ fixed-length segments:
\begin{equation}
    \hat{\mathbf{x}}^{(1)}_{\mathrm{seg}, i}
        = \hat{\mathbf{x}}^{(1)}_{t+s_i : t+e_i},
    \qquad i = 1, \ldots, N,
\end{equation}
where $[s_i, e_i]$ denotes the index range of the $i$-th segment extracted from the predicted future. Each segment is appended to the original input window to form an augmented input:
\begin{equation}
    \mathbf{x}^{(i)}_{\mathrm{aug}}
        = \left[\,\mathbf{x}_{t-L+1:t} \;;\;
        \hat{\mathbf{x}}^{(1)}_{\mathrm{seg}, i}\right],
    \qquad i = 1, \ldots, N.
\end{equation}
This yields $N$ inputs that share the same historical window but differ in the attached future reference segment, which are then used in the parallel self-refinement stage.

\subsection{Parallel Self-Refinement}
Given the augmented input sequences, we train $N$ independent second-stage predictors. For each augmented input $\mathbf{x}^{(i)}_{\mathrm{aug}}$, a separate model $f^{(i)}_{\phi}$ predicts the same $H$-step horizon:
\begin{equation}
    \hat{\mathbf{x}}^{(2)}_{i}
        = f^{(i)}_{\phi}\!\left(\mathbf{x}^{(i)}_{\mathrm{aug}}\right),
    \qquad i = 1, \ldots, N,
\end{equation}
where each model is trained to minimize the forecasting error:
\begin{equation}
    \phi^{*}_{i}
        = \arg\min_{\phi_{i}}
        \mathcal{L}
        \Big(
            f^{(i)}_{\phi}\!\left(\mathbf{x}^{(i)}_{\mathrm{aug}}\right),
            \mathbf{x}_{t+1:t+H}
        \Big).
\end{equation}

Because each model receives a different future segment as augmented context, it learns a distinct refinement pattern for the first-stage forecast. Although no explicit residual model is introduced, the augmented inputs naturally induce implicit corrections:
\begin{equation}
    \hat{\mathbf{x}}^{(2)}_{i}
        \;\approx\;
        \hat{\mathbf{x}}^{(1)}
        + \text{\textit{segment-specific adjustment}}.
\end{equation}
Thus, all models share the same historical window but learn different future-aligned refinement directions from their segment.

\subsection{Ensembled Forecasting}
To obtain a stable final forecast, we aggregate the $N$ second-stage predictions using a step-wise top-$K$ ensemble. Let the second-stage predictions be ranked by their validation performances as
\begin{equation}
    \hat{\mathbf{x}}^{(2)}_{1st},\;
    \hat{\mathbf{x}}^{(2)}_{2nd},\;
    \ldots,\;
    \hat{\mathbf{x}}^{(2)}_{Nth},
\end{equation}
where $1st$ denotes the best-performing model. For each candidate ensemble size $K \in \{M,\, 2M,\, \ldots,\, N\}$ with a step increment $M$,
we compute the ensemble prediction by averaging the top-$K$ models:
\begin{equation}
    \hat{\mathbf{x}}^{(K)}
        = \frac{1}{K}
          \sum_{i=1}^{K}
          \hat{\mathbf{x}}^{(2)}_{i\text{-}th}.
\end{equation}
This step-wise expansion allows us to explore ensembles of increasing diversity without an exhaustive combination search. 

However, including diverse models in a top-$K$ ensemble does not always improve accuracy; a large $K$ may introduce unstable or redundant predictors, whereas a small $K$ provides limited variance reduction. Thus, to select an optimal ensemble size $K^{*}$, we revisit the ensemble error decomposition by Ueda and Nakano~\citep{ensemble}:
\begin{equation}
\small
    \mathbb{E}\big[(y - \hat{y}_{\mathrm{ens}})^{2}\big]
    = \mathrm{Bias}^{2}
    + \frac{1}{K}\,\mathrm{Var}
    + \frac{2}{K(K-1)}
      \sum_{i<j} \mathrm{Cov}(\varepsilon_i, \varepsilon_j)
    + \sigma_{\mathrm{noise}}^{2}.
\end{equation}
This indicates that a good ensemble requires \textit{low prediction variance} and \textit{low inter-model covariance}.
As both the variance and covariance terms depend on the error values $\varepsilon_i$, which are unavailable at inference time, we approximate them using \emph{observable} statistics of predictions:
\begin{equation}
\small
    V(K)
    = \mathrm{Var}\!\left(
        \hat{\mathbf{x}}^{(2)}_{1},
        \ldots,
        \hat{\mathbf{x}}^{(2)}_{K}
      \right) \text{ and } 
          R(K)
    = \mathrm{MeanCorr}\!\left(
        \hat{\mathbf{x}}^{(2)}_{1},
        \ldots,
        \hat{\mathbf{x}}^{(2)}_{K}
      \right).
\end{equation}

These statistics serve as practical surrogates for the variance and covariance terms in the decomposition. They also help detect outlier predictors (high $V(K)$) and redundant predictors (high $R(K)$).
Alternatively, we adopt a min--max normalized formulation:
\begin{equation}
    S(K)
    =
    \frac{V(K) - V_{\min}}{V_{\max} - V_{\min} + \varepsilon}
    +
    \frac{R(K) - R_{\min}}{R_{\max} - R_{\min} + \varepsilon},
\end{equation}
which provides scale-invariant normalization and prevents degeneracy via a small $\varepsilon > 0$. The optimal $K$ is then chosen as
\begin{equation}
    K^{*} = \arg\min_{K} S(K).
\end{equation}

\section{Experiments}

\subsection{Experiment Settings}
We evaluate BTTF on four real-world benchmarks (ETTh1, ETTm2, Exchange Rate, and ILI) under the univariate forecasting setting. 
We adopt linear-based models, Linear and DLinear, as base models using the official implementation provided by the LTSF-Linear repository~\citep{dlinear}. We choose two recent methods, iTransformer and SegRNN, from the Time-Series-Library benchmark~\citep{experimentbaseline} for comparison with advanced architectures. We vary the forecasting horizons for each dataset and measure accuracy in terms of MSE (mean squared error) and MAE (mean absolute error).

For fair comparison, all first- and second-stage models in BTTF utilize identical hyperparameters and are trained using either early stopping or a single-epoch setting.
The window size for segmentation is fixed to one-third of the prediction horizon, with a stride of 1 for ILI and \{1, 2, 4, 8\} for the other datasets. In the step-wise ensemble, the step size is fixed to $M = 5$. The implementation details and source codes are provided at \url{https://github.com/OSsunNO/BTTF}.

\subsection{Overall Performance}
The key findings of the evaluation results in Table~\ref{tab:overall_performance} are:

\begin{itemize}[leftmargin=9pt, noitemsep]
\item Applying BTTF to simple Linear and DLinear consistently improves accuracy in most settings, and in many cases achieves competitive or even superior performance compared to strong baselines such as iTransformer and SegRNN. 
\item Even when the first-stage model is poorly optimized due to unstable or insufficient training, the second-stage models often surpass the best performance reported for the base model, achieving improvements of up to 35\%.
\item The simplest 1E--1E setting still yields strong results, highlighting that the step-wise ensemble efficiently leverages diverse models. 
\end{itemize}

On ETTm2, however, where the first-stage model is sufficiently fitted, BTTF yields a slight performance drop (around 2\% on average). This suggests that our framework can also serve as a practical diagnostic tool to identify whether a base model is fully optimized and has little room for further improvement. Alternatively, a more advanced ensemble strategy could further enhance performance as our strategy conservatively excludes worst-case predictors. 


\twocolumn[
\begin{center}
\begin{minipage}{\textwidth}
\centering
\setlength{\tabcolsep}{3pt}

\captionof{table}{
Forecasting accuracy results in MSE and MAE over different horizons. Training strategies of BTTF applied for the first- or second-stage model are denoted as ES (early stopping) or 1E (1 epoch), respectively (e.g., ES-1E means early stopping for the first stage and 1 epoch for the second stage). Performance gain over the base model by BTTF is given in parentheses (\%). Best results are in bold, and second-best results are underlined. 
}

\label{tab:overall_performance}

\begin{adjustbox}{max width=\textwidth}
\begin{tabular}{
l|c|
cc|>{\columncolor{gray!15}}c>{\columncolor{gray!15}}c|>{\columncolor{gray!15}}c>{\columncolor{gray!15}}c|cc|>{\columncolor{gray!15}}c>{\columncolor{gray!15}}c|>{\columncolor{gray!15}}c>{\columncolor{gray!15}}c|cc|cc}
\toprule
\textbf{Dataset} & \textbf{Horizon} &
\multicolumn{2}{c|}{\textbf{Linear}} &
\multicolumn{2}{c|}{\cellcolor{gray!15}\textbf{Linear+BTTF (ES-1E)}} &
\multicolumn{2}{c|}{\textbf{\cellcolor{gray!15}Linear+BTTF (1E-1E)}} &
\multicolumn{2}{c|}{\textbf{DLinear}} &
\multicolumn{2}{c|}{\textbf{\cellcolor{gray!15}DLinear+BTTF (ES-1E)}} &
\multicolumn{2}{c|}{\textbf{\cellcolor{gray!15}DLinear+BTTF (1E-1E)}} &
\multicolumn{2}{c|}{\textbf{iTransformer}} &
\multicolumn{2}{c}{\textbf{SegRNN}} \\

 & &
\textbf{MSE} & \textbf{MAE} &
\textbf{MSE (\%)} & \textbf{MAE (\%)} &
\textbf{MSE (\%)} & \textbf{MAE (\%)} &
\textbf{MSE} & \textbf{MAE} &
\textbf{MSE (\%)} & \textbf{MAE (\%)} &
\textbf{MSE (\%)} & \textbf{MAE (\%)} &
\textbf{MSE} & \textbf{MAE} &
\textbf{MSE} & \textbf{MAE}  \\
\midrule

\textbf{ILI} & 24 & 1.6197 & 1.1315 & 0.7097 (56.2$\textcolor{red}{\blacktriangle}$) & 0.6703 (40.8$\textcolor{red}{\blacktriangle}$) & 0.6682 (58.7$\textcolor{red}{\blacktriangle}$) & 0.6547 (42.1$\textcolor{red}{\blacktriangle}$) & 0.7035 & 0.6638 & \underline{0.6598 (6.2$\textcolor{red}{\blacktriangle}$)} & \textbf{0.6124 (7.8$\textcolor{red}{\blacktriangle}$)} & \textbf{0.6475 (8.0$\textcolor{red}{\blacktriangle}$)} & \underline{0.6189 (6.8$\textcolor{red}{\blacktriangle}$)} & 0.6797 & 0.6439 & 0.7244 & 0.6280  \\
                 
                 & 36 & 0.8632 & 0.8048 & \textbf{0.6837 (20.8$\textcolor{red}{\blacktriangle}$)} & \underline{0.6669 (17.1$\textcolor{red}{\blacktriangle}$)} & \underline{0.6902 (20.1$\textcolor{red}{\blacktriangle}$)} & 0.6725 (16.4$\textcolor{red}{\blacktriangle}$) & 0.7943 & 0.7403 & 0.7019 (11.6$\textcolor{red}{\blacktriangle}$) & \textbf{0.6505 (12.1$\textcolor{red}{\blacktriangle}$)} & 0.7588 (4.5$\textcolor{red}{\blacktriangle}$) & 0.7070 (4.5$\textcolor{red}{\blacktriangle}$) & 0.7492 & 0.7009 & 0.8678 & 0.7119  \\
                 
                 & 48 & 0.8469 & 0.8011 & \textbf{0.7180 (15.2$\textcolor{red}{\blacktriangle}$)} & \textbf{0.6895 (13.9$\textcolor{red}{\blacktriangle}$)} & \underline{0.7233 (14.6$\textcolor{red}{\blacktriangle}$)} & \underline{0.6970 (13.0$\textcolor{red}{\blacktriangle}$)} & 0.8615 & 0.7931 & 0.7861 (8.8$\textcolor{red}{\blacktriangle}$) & 0.7174 (9.6$\textcolor{red}{\blacktriangle}$) & 0.7686 (10.8$\textcolor{red}{\blacktriangle}$) & 0.7266 (8.4$\textcolor{red}{\blacktriangle}$) & 0.8123 & 0.7450 & 0.9421 & 0.7596 \\
                 
                 & 60 & 0.8783 & 0.7901 & \textbf{0.7464 (15.0$\textcolor{red}{\blacktriangle}$)} & \textbf{0.7124 (9.8$\textcolor{red}{\blacktriangle}$)} & 0.8246 (6.1$\textcolor{red}{\blacktriangle}$) & 0.7601 (3.8$\textcolor{red}{\blacktriangle}$) & 0.9693 & 0.8668 & \underline{0.7956 (17.9$\textcolor{red}{\blacktriangle}$)} & \underline{0.7405 (14.6$\textcolor{red}{\blacktriangle}$)} & 0.8826 (9.0$\textcolor{red}{\blacktriangle}$) & 0.8046 (7.2$\textcolor{red}{\blacktriangle}$)  & 0.9494 & 0.8205 & 0.9268 & 0.7768 \\
\midrule

\textbf{Exchange Rate} & 96 & 0.1118 & 0.2555 & 0.1043 (6.8$\textcolor{red}{\blacktriangle}$) & 0.2455 (3.9$\textcolor{red}{\blacktriangle}$) & 0.1021 (8.7$\textcolor{red}{\blacktriangle}$) & 0.2458 (3.8$\textcolor{red}{\blacktriangle}$) & 0.1053 & 0.2495 & 0.1020 (3.2$\textcolor{red}{\blacktriangle}$)  & 0.2401 (3.8$\textcolor{red}{\blacktriangle}$) & \underline{0.0991 (5.9$\textcolor{red}{\blacktriangle}$)} & 0.2415 (3.2$\textcolor{red}{\blacktriangle}$)  & 0.1024 & \underline{0.2372} & \textbf{0.0966} & \textbf{0.2279}  \\
                 
                 & 192 & 0.2473 & 0.3811 & 0.2079 (16.0$\textcolor{red}{\blacktriangle}$) & 0.3539 (7.1$\textcolor{red}{\blacktriangle}$) & 0.2333 (5.7$\textcolor{red}{\blacktriangle}$) & 0.3827 (0.4$\textcolor{blue}{\blacktriangledown}$) & 0.3080 & 0.4067 & 0.2159 (29.9$\textcolor{red}{\blacktriangle}$) & 0.3552 (12.7$\textcolor{red}{\blacktriangle}$) & \textbf{0.2001 (35.0$\textcolor{red}{\blacktriangle}$)}  & 0.3535 (13.1$\textcolor{red}{\blacktriangle}$)  & 0.2075 & \underline{0.3416} & \underline{0.2034} & \textbf{0.3353}  \\
                 
                 & 336 & 0.4276 & 0.5129 & \underline{0.3091 (27.7$\textcolor{red}{\blacktriangle}$)} & \underline{0.4571 (10.9$\textcolor{red}{\blacktriangle}$)} & 0.4158 (2.8$\textcolor{red}{\blacktriangle}$) & 0.5071 (1.1$\textcolor{red}{\blacktriangle}$) & 0.3339 & 0.4766 & 0.3339 (-) & 0.4668 (2.6$\textcolor{red}{\blacktriangle}$) & \textbf{0.3083 (7.7$\textcolor{red}{\blacktriangle}$)} & \textbf{0.4544 (4.7$\textcolor{red}{\blacktriangle}$)}  & 0.4190 & 0.4832 & 0.4311 & 0.5001  \\
                 
                 & 720 & 1.1591 & 0.8464 & \underline{0.8698 (25.0$\textcolor{red}{\blacktriangle}$)} & 0.7370 (12.9$\textcolor{red}{\blacktriangle}$) & 0.9278 (20.0$\textcolor{red}{\blacktriangle}$) & 0.7596 (10.3$\textcolor{red}{\blacktriangle}$) & 1.2503 & 0.8795 & 0.8760 (29.9$\textcolor{red}{\blacktriangle}$) & \underline{0.7281 (17.7$\textcolor{red}{\blacktriangle}$)} & \textbf{0.8589 (31.3$\textcolor{red}{\blacktriangle}$)} & \textbf{0.7232 (17.8$\textcolor{red}{\blacktriangle}$)} & 1.1050 & 0.8060 & 1.2795 & 0.8724  \\
                 
\midrule
\textbf{ETTh1} & 96  & 0.0586 & 0.1833 & 0.0559 (4.6$\textcolor{red}{\blacktriangle}$) & 0.1798 (1.9$\textcolor{red}{\blacktriangle}$) & 0.0554 (5.5$\textcolor{red}{\blacktriangle}$) & 0.1797 (2.0$\textcolor{red}{\blacktriangle}$) & 0.0580  & 0.1805 & \textbf{0.0543 (6.3$\textcolor{red}{\blacktriangle}$)} & \textbf{0.1778 (1.5$\textcolor{red}{\blacktriangle}$)} & \underline{0.0553 (4.6$\textcolor{red}{\blacktriangle}$)} & \underline{0.1781 (1.4$\textcolor{red}{\blacktriangle}$)}  & 0.0576 & 0.1845 & 0.0555 & 0.1838  \\
               
               & 192 & 0.0768 & 0.2112 & 0.0737 (4.0$\textcolor{red}{\blacktriangle}$)  & 0.2103 (0.4$\textcolor{red}{\blacktriangle}$) & 0.0763 (0.7$\textcolor{red}{\blacktriangle}$) & 0.2094 (0.9$\textcolor{red}{\blacktriangle}$) & \underline{0.0727} & 0.2126 & \textbf{0.0717 (1.4$\textcolor{red}{\blacktriangle}$)} & \underline{0.2061 (3.1$\textcolor{red}{\blacktriangle}$)} & 0.0729 (0.3$\textcolor{blue}{\blacktriangledown}$)  & 0.2066 (2.8$\textcolor{red}{\blacktriangle}$)  & \underline{0.0727} & \textbf{0.2059} & 0.0757 & 0.2182 \\
               
               & 336 & 0.0976 & 0.2443 & \underline{0.0867 (11.3$\textcolor{red}{\blacktriangle}$)} & \underline{0.2337 (4.4$\textcolor{red}{\blacktriangle}$)} & 0.0921 (5.7$\textcolor{red}{\blacktriangle}$) & 0.2383 (2.5$\textcolor{red}{\blacktriangle}$) & 0.0997 & 0.2464 & 0.0934 (6.3$\textcolor{red}{\blacktriangle}$) & 0.2393 (2.9$\textcolor{red}{\blacktriangle}$) & 0.0924 (7.3$\textcolor{red}{\blacktriangle}$) & 0.2391 (3.0$\textcolor{red}{\blacktriangle}$) & \textbf{0.0829} & \textbf{0.2217} & 0.0905 & 0.2399 \\
               
               & 720 & 0.1598 & 0.3257 & 0.1535 (4.0$\textcolor{red}{\blacktriangle}$) & 0.3175 (2.5$\textcolor{red}{\blacktriangle}$) & 0.1665 (4.1$\textcolor{blue}{\blacktriangledown}$) & 0.3322 (2.0$\textcolor{blue}{\blacktriangledown}$) & 0.1866 & 0.3570 & 0.1669 (10.6$\textcolor{red}{\blacktriangle}$) & 0.3304 (7.5$\textcolor{red}{\blacktriangle}$) & 0.1749 (6.3$\textcolor{red}{\blacktriangle}$) & 0.3444 (3.5$\textcolor{red}{\blacktriangle}$)  & \underline{0.0807} & \textbf{0.2247} & \textbf{0.0781} & \underline{0.2257}  \\
\midrule

\textbf{ETTm2} & 96    & 0.0662 & 0.1894 & 0.0662 (-) & 0.1898 (0.2$\textcolor{blue}{\blacktriangledown}$) & 0.0674 (1.7$\textcolor{blue}{\blacktriangledown}$) & 0.1917 (1.2$\textcolor{blue}{\blacktriangledown}$) & \textbf{0.0638} &  \textbf{0.1841} & \underline{0.0639 (0.2$\textcolor{blue}{\blacktriangledown}$)} & 0.1845 (0.2$\textcolor{blue}{\blacktriangledown}$) & 0.0640 (0.3$\textcolor{blue}{\blacktriangledown}$) & \underline{0.1843 (0.1$\textcolor{blue}{\blacktriangledown}$)} & 0.0684 & 0.1883 & 0.0686 & 0.1891  \\
                       
                       & 192 & 0.0935 & 0.2299 & 0.0945 (1.0$\textcolor{blue}{\blacktriangledown}$) & 0.2299 (0.6$\textcolor{blue}{\blacktriangledown}$) & 0.0941 (0.6$\textcolor{blue}{\blacktriangledown}$) & 0.2299 (0.5$\textcolor{blue}{\blacktriangledown}$) & \textbf{0.0914} & \textbf{0.2259} & \underline{0.0921 (0.8$\textcolor{blue}{\blacktriangledown}$)} & \underline{0.2271 (0.5$\textcolor{blue}{\blacktriangledown}$)} & 0.0924 (1.2$\textcolor{blue}{\blacktriangledown}$) & 0.2275 (0.7$\textcolor{blue}{\blacktriangledown}$)  & 0.1090 & 0.2465 & 0.1027 & 0.2377  \\
                       
                       & 336 & 0.1233 & 0.2658 & 0.1335 (8.3$\textcolor{blue}{\blacktriangledown}$)& 0.2808 (5.6$\textcolor{blue}{\blacktriangledown}$) & \underline{0.1211 (1.8$\textcolor{red}{\blacktriangle}$)} & \underline{0.2642 (0.6$\textcolor{red}{\blacktriangle}$)} & \textbf{0.1189} & \textbf{0.2613} & 0.1225 (3.0$\textcolor{blue}{\blacktriangledown}$) & 0.2666 (2.0$\textcolor{blue}{\blacktriangledown}$) & 0.1243 (4.5$\textcolor{blue}{\blacktriangledown}$) & 0.2673 (2.3$\textcolor{blue}{\blacktriangledown}$) & 0.1425 & 0.2869 & 0.1335 & 0.2783  \\
                       
                       & 720 & \underline{0.1771} & \underline{0.3225} & 0.1819 (2.8$\textcolor{blue}{\blacktriangledown}$) & 0.3276 (1.6$\textcolor{blue}{\blacktriangledown}$) & 0.1798 (1.6$\textcolor{blue}{\blacktriangledown}$) & 0.3274 (1.5$\textcolor{blue}{\blacktriangledown}$) & \textbf{0.1740} & \textbf{0.3201} & 0.1856 (6.7$\textcolor{blue}{\blacktriangledown}$) & 0.3360 (5.0$\textcolor{blue}{\blacktriangledown}$) & 0.1866 (7.2$\textcolor{blue}{\blacktriangledown}$) & 0.3321 (3.8$\textcolor{blue}{\blacktriangledown}$) & 0.1850 & 0.3370 & 0.1858 & 0.3347 \\

\bottomrule
\end{tabular}
\end{adjustbox}
\end{minipage}
\end{center}
\vspace{2em}
]
\noindent

\subsection{Computational Efficiency}
Although BTTF employs multiple augmented models, the additional computational cost remains manageable. Because the first-stage model is identical to the base model, its training time is unchanged. The second stage introduces extra overhead, requiring approximately 8.3$\times$ training time on average; however, this gap is reduced relative to advanced baselines, yielding an average of 2.6$\times$ processing time compared to iTransformer and SegRNN, or even a 21\% decrease on the ILI dataset. Furthermore, as the second-stage models are independent, this overhead can be substantially reduced by distributing them across multiple processing units for parallel training. The second-stage models also share the same historical input sequence and differ only in the appended augmentation segment, contributing to a compact memory footprint during training.




\section{Conclusion and Future Work}
This work introduced BTTF, a simple framework that leverages forecasting-based augmentation and self-refinement for LTSF. While existing studies mainly focus on architectural innovations to model input-level interactions, our approach shows that prediction performance can be further enhanced by using a model's own future predictions as augmentation signals. 
Experimental results demonstrate that even lightweight linear models benefit substantially from BTTF, achieving up to 58\% performance improvement. These results suggest that the future context of an initial prediction can serve as an effective additional feature. 
Overall, our framework adopts the benefits of both DMS and IMS by using forecasts not only as outputs but also as informative context for refining predictions, opening new perspectives for LTSF.


For future work, we will extend BTTF to multivariate forecasting by leveraging future predictions as both temporal context and cross-variable signals, while also incorporating predictions from heterogeneous model architectures to enable refinement based on diverse inductive biases. We further plan to explore adaptive segment selection and refinement strategies to better align the framework with dataset-specific temporal characteristics.

\newpage




\section*{Acknowledgments}
This work was partly supported by the Institute of Information \& Communications Technology Planning \& Evaluation (IITP)-ICT Creative Consilience Program (IITP-2026-RS-2020-II201819), Information Technology Research Center (ITRC) (IITP-2026-RS-2024-00436857), Artificial Intelligence Star Fellowship Program (IITP-2026-RS-2025-02304828) funded by the Korea government(MSIT), and the Korea Institutes of Police Technology (KIPoT) (RS-2025-02218280) funded by the Korean National Police Agency.

\balance
\bibliographystyle{ACM-Reference-Format}
\bibliography{References}

@inproceedings{traffic,
  title={Traffic flow prediction via spatial temporal graph neural network},
  author={Wang, Xiaoyang and Ma, Yao and Wang, Yiqi and Jin, Wei and Wang, Xin and Tang, Jiliang and Jia, Caiyan and Yu, Jian},
  booktitle={Proceedings of the Web conference 2020},
  year={2020}
}

@inproceedings{rest,
  title={Rest: Relational event-driven stock trend forecasting},
  author={Xu, Wentao and Liu, Weiqing and Xu, Chang and Bian, Jiang and Yin, Jian and Liu, Tie-Yan},
  booktitle={Proceedings of the Web conference 2021},
  year={2021}
}

@inproceedings{bigforecasting,
  title={Forecasting big time series: Theory and practice},
  author={Faloutsos, Christos and Flunkert, Valentin and Gasthaus, Jan and Januschowski, Tim and Wang, Yuyang},
  booktitle={Proceedings of the 25th ACM SIGKDD International Conference on Knowledge Discovery \& Data Mining},
  year={2019}
}

@inproceedings{dlinear,
  title={Are transformers effective for time series forecasting?},
  author={Zeng, Ailing and Chen, Muxi and Zhang, Lei and Xu, Qiang},
  booktitle={Proceedings of the AAAI conference on artificial intelligence},
  year={2023}
}

@inproceedings{itransformer,
  title={iTransformer: Inverted transformers are effective for time series forecasting},
  author={Liu, Yong and Hu, Tengge and Zhang, Haoran and Wu, Haixu and Wang, Shiyu and Ma, Lintao and Long, Mingsheng},
  booktitle={The Twelfth International Conference on Learning Representations},
  year={2024}
}

@article{segrnn,
  title={Seg{RNN}: Segment recurrent neural network for long-term time series forecasting},
  author={Lin, Shengsheng and Lin, Weiwei and Wu, Wentai and Zhao, Feiyu and Mo, Ruichao and Zhang, Haotong},
  journal={IEEE Internet of Things Journal},
  year={2025},
  publisher={IEEE}
}

@inproceedings{wentime,
  title={Time series data augmentation for deep learning: a survey},
  author={Wen, Qingsong and Sun, Liang and Yang, Fan and Song, Xiaomin and Gao, Jingkun and Wang, Xue and Xu, Huan},
  booktitle={Proceedings of the Thirtieth International Joint Conference on Artificial Intelligence},
  year={2021}
}

@inproceedings{demirel2023finding,
  title={Finding order in chaos: A novel data augmentation method for time series in contrastive learning},
  author={Demirel, Berken Utku and Holz, Christian},
  journal={Advances in Neural Information Processing Systems},
  year={2023}
}

@inproceedings{staug,
  title={Towards diverse and coherent augmentation for time-series forecasting},
  author={Zhang, Xiyuan and Chowdhury, Ranak Roy and Shang, Jingbo and Gupta, Rajesh and Hong, Dezhi},
  booktitle={IEEE International Conference on Acoustics, Speech and Signal Processing},
  year={2023}
}

@inproceedings{ensemble,
  title={Generalization error of ensemble estimators},
  author={Ueda, Naonori and Nakano, Ryohei},
  booktitle={Proceedings of International Conference on Neural Networks},
  year={1996},
}

@inproceedings{experimentbaseline,
  title={Deep time series models: a comprehensive survey and benchmark},
  author={Yuxuan Wang and Haixu Wu and Jiaxiang Dong and Yong Liu and Mingsheng Long and Jianmin Wang},
  booktitle={arXiv preprint arXiv:2407.13278},
  year={2024}
}

\end{document}